# Search-based optimal motion planning for automated driving

Zlatan Ajanovic[1,3], Bakir Lacevic[2], Barys Shyrokau[3], Michael Stolz[1] and Martin Horn[4]

*Abstract*— This paper presents a framework for fast and robust motion planning designed to facilitate automated driving. The framework allows for real-time computation even for horizons of several hundred meters and thus enabling automated driving in urban conditions. This is achieved through several features. Firstly, a convenient geometrical representation of both the search space and driving constraints enables the use of classical path planning approach. Thus, a wide variety of constraints can be tackled simultaneously (other vehicles, traffic lights, etc.). Secondly, an exact cost-to-go map, obtained by solving a relaxed problem, is then used by A*-based algorithm with model predictive flavour in order to compute the optimal motion trajectory. The algorithm takes into account both distance and time horizons. The approach is validated within a simulation study with realistic traffic scenarios. We demonstrate the capability of the algorithm to devise plans both in fast and slow driving conditions, even when full stop is required.

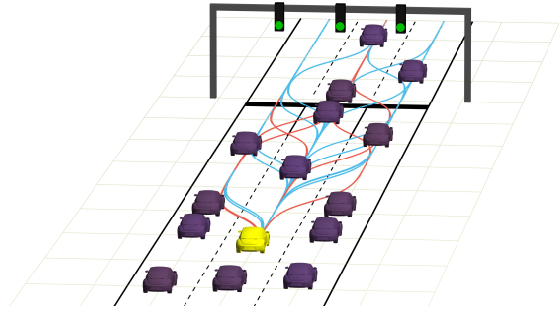

Fig. 1: Ilustration of demonstrated driving scenario consisting of multilane driving in presence of traffic lights.

## I. INTRODUCTION

Automated driving promises to significantly reduce the number of road fatalities, increase traffic efficiency, reduce environmental pollution, give mobility to handicapped members of our society and empower new services such as mobility on demand [1]. If these promises are fulfilled automated driving could eventually change mobility.

Vehicle automation is based on classic robotics Sensing-Planning-Acting cycle, where Motion Planning (MP) is the crucial step. Task of MP is to provide a collision-free motion plan from the given starting pose to the given goal region, taking into account system dynamics, obstacles and possibly desired criteria (cost function). MP has been researched since 1970s [2], mostly in robotics. However, vehicle automation application brings new challenges as the environment is cluttered, dynamic, complex, uncertain and the vehicle is often operating on the limits of its dynamics. Several works give a comprehensive overview of current motion planning approaches in vehicle automation domain [3]–[5]. Usually, MP for automated vehicles is structured hierarchically [3], with route planning at the top, operating with the smallest frequency (e.g. once a trip), followed by the behavioral layer responsible for making a decision on which maneuver should be executed. When a decision is known, the local MP layer generates a trajectory or a waypoint that satisfies safety and traffic rules, and is further executed by a stabilizing controller (benchmarking of different stabilizing controllers for path-following is presented in [6]).

*https://youtu.be/D5XJ5ncSuqQ
[1]Z. Ajanovic and M. Stolz are with Virtual Vehicle Research Center, Inffeldgasse 21a, 8010 Graz, Austria. {zlatan.ajanovic, michael.stolz}@v2c2.at
[2]B. Lacevic is with the Faculty of Electrical Engineering, University of Sarajevo, 7100 Sarajevo, Bosnia and Herzegovina. bakir.lacevic@etf.unsa.ba
[3]B. Shyrokau and Z. Ajanovic are with the Department of Cognitive Robotics, Delft University of Technology, The Netherlands, Mekelweg 2, 2628 CD Delft, The Netherlands. b.shyrokau@tudelft.nl
[4]M. Horn is head of Institute of Automation and Control at Graz University of Technology, Inffeldgasse 21b, 8010 Graz, Austria. martin.horn@tugraz.at

The behavioral layer was initially implemented using finite state machines, and most of the participants in DARPA Urban challenge used it [7]–[9]. To deal with uncertainty, solutions based on Partially Observable Markov Decision Process (POMDP) [10] were also proposed. In general, decision making requires a sample trajectory to estimate whether a certain maneuver is possible. This natural coupling of trajectory planning and decision making calls for the integration of behavioral and local MP layer [5]. However, integration of the behavioral and local MP layer introduces combinatorial aspects into the driving problem [11].

Multi-lane driving problem has been tackled by several approaches with limited success. In [12], authors proposed spatiotemporal state lattices used with a dynamic programming search to plan collision-free motion in the presence of dynamic obstacles. The proposed search was rather fast (less than 20 ms), yet only a limited number (7) of velocity variants were used and lattice construction is such that full stop is not possible. In [13], authors formalized the generation of all possible combinations and used local planning [11] for each one of them. The best of them is chosen as the global optimal result. This approach is not applicable to environments where many combinations are possible, especially where traffic lights are present as they introduce infinitively many combinations. Several authors [14], [15] used mixed integer programming approaches to treat multiple variants with the assumption that the desired velocity is defined and the deviation from this velocity is used within a cost function for optimization. This simplifica-tion leads to the local optimal solution as the gap for the lane change can be influenced by the velocity history of the ego vehicle. An interesting approach for MP without considering the driving rules was presented in [16] where the authors used heuristic search to plan the ego vehicle motion in the dynamic environment. None of the mentioned approaches considers traffic lights that represent a common constraint within urban driving.

On the other hand, MP in the presence of traffic lights has been widely studied with the aim at improving energy efficiency and reducing trip time. The approach is usually hierarchical. On the top level, kinematic limits on velocity or desired velocity are determined such that the vehicle can pass one or more traffic lights without stopping. The output is then fed to MPC-based local motion planning. The approaches for top level planning vary, from a simple kinematics [17] to Dijkstra's algorithm [18] and supervisory MPC [19]. Local motion planning is usually based on MPC, which may consider other vehicles as well, but only vehicle following and single-lane driving [17], [19] is considered. To the best of authors' knowledge, none of the related work tackles combined multi-lane driving in the presence of traffic lights.

The framework proposed in this paper relies on search algorithm in MPC-like scheme and reusing the cost-to-go map to increase the search efficiency. A conceptually similar approach can be found in [9], where authors used the cost-to-go map (from the global path planning) with a MPC-like forward trajectory planner to achieve the shortest time travel. In the [20], the authors presented the planner for mobile vehicle operating on poorly traversable terrains. In the [21], authors presented a similar approach using forward dynamic programming in MPC scheme for energy optimal MP. Moreover, the framework proposed in this paper relies on search-based planning algorithms (within MPC scheme). Probably the most notable uses of search algorithms for automated driving are for planning in unstructured environment in DARPA Urban challenge [22], [23] and for velocity planning in Prometheus project [24].

The main contribution of this work is a comprehensive motion planning framework for automated driving that includes the following features:
- a convenient search space definition, enabling intuitive formulation of a wide variety of constraints,
- the possibility to reuse backward planning results from a relaxed problem for shorter planning times,
- integrated reference lane decision making and velocity trajectory planning (longitudinal and lateral motion),
- hybrid time/distance horizon and discretization steps that enable both slow and fast trajectories,
- search in continuous time, distance and lane space provided by hybrid A* search,
- linear lateral motion model for efficient and effective lane-change planning,
- a novel demonstration on the complex use-case with multi-lane driving in a presence of traffic lights.

The remainder of this paper is structured as follows. In section II, we define a problem of driving by presenting the vehicle model, a defining search space and obstacles. In section III, we describe the optimal MP framework. In section IV, the results of simulation study are presented, followed by the concluding remarks in section V.

## II. PROBLEM STATEMENT

In this section, we define the addressed problem. First, a general driving problem is defined, followed by the formulation of search space and obstacles. Finally, we provide the details about the vehicle model and the cost evaluation.

### A. Driving in a complex, dynamic environment

Driving is a complex task consisting of continuous planning and execution in order to achieve desired goals and avoid collisions with other participants, obey traffic rules, comply with vehicle dynamics and factors like comfort, safety and efficiency. To fully automate driving, the vehicle has to be able to autonomously make decisions and plan its motion, while considering all mentioned requirements. The environment is usually highly dynamic, with speeds that may reach $50m/s$ or above. Moreover, the environment is complex, including many different participants, traffic rules, traffic control devices, etc. The mentioned conditions impose many different constraints on the driving trajectory [25].

The mayor challenges can be stated as: i) vehicle dynamics influence feasibility of the plans, ii) dynamic constraints are not known during initial planning, iii) the real motion of other participants deviates from the predicted one, iv) planning for long horizons with dynamic constraints is computationally expensive, v) long horizon planning is necessary to achieve long-term benefits such as energy efficiency, vi) conservative assumptions narrow down the search space, which can cause the loss of solution even in the case it exists. Therefore, a frequent replanning with long horizons is necessary while considering the vehicle model and environment as well.

### B. Search space

To tackle dynamic obstacles and avoid the risk of losing quality solutions, we use the 3D space-time $\Omega$ as a search space via Cartesian product of 2D configuration space and time dimension [24]:

$$\Omega = \left\{ \mathbf{q} \equiv [t, s, l]^T \mid t \in \mathbb{R}^+, s \in \mathbb{R}^+, l \in [1, N_l] \right\}. \quad (1)$$

Here, $t$ is time, $s$ is the longitudinal position along the road and $l$ is the lateral position on a road. Dimension $l$ is defined such that the middle of the rightmost lane has value 1 while the middle of the left-most lane has value $N_l$. The value $1.5$ means that vehicle is halfway between lane 1 and 2.

### C. Obstacles

We consider several types of constraints/obstacles, such as constraints imposed by other vehicles, traffic lights, speed limits and forbidden lane-change.

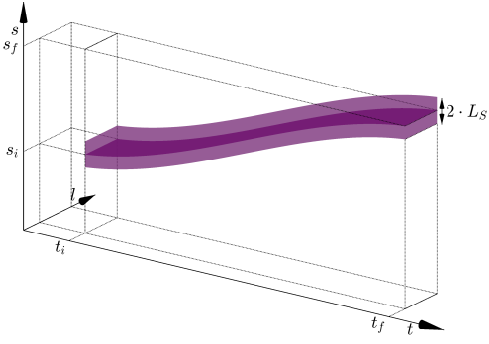

Fig. 2: Obstacle created by vehicle which is speeding up and slowing down.

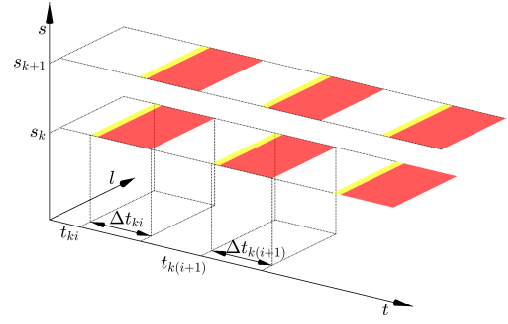

Fig. 3: Obstacles created by two traffic lights.

*1) Vehicle obstacle:* Other vehicles on the road represent obstacles for ego vehicle and constrain it's motion. The violation of these constraints can be manifested not only as a direct collision, but sometimes also as a violation of the driving rules, such as overtaking from right side or slow overtaking from the left side. For a certain vehicle $\mathcal{V}_k$, the trajectory of its center is described with $s_k(t)$ and $l_k(t)$, while suitable lower and upper bounds can be defined as:

$$L_S = L_k/2 + L_{ego}/2, \quad (2a)$$
$$\underline{s}_k(t) = s_k(t) - L_S, \quad (2b)$$
$$\overline{s}_k(t) = s_k(t) + L_S, \quad (2c)$$

where $L_k$ is the length of vehicle $\mathcal{V}_k$ and $L_{ego}$ is the length of ego vehicle. Thus, the $L_{ego}$ is practically incorporated within the obstacle so ego vehicle can be considered as a point. Based on this, the corresponding obstacle can be defined as:

$$\mathcal{O}_k^V = \{\mathbf{q} \in \Omega \mid s(t) \in [\underline{s}_k(t), \overline{s}_k(t)]\}. \quad (3)$$

A collision check for a given $\mathbf{q}$, or the condition for which collision occurs can be validated by:

$$\tau\left(\exists \mathbf{q} \in \mathcal{O}_k^V \mid l(t) \in (l_k(t) - 1, l_k(t) + 1)\right), \quad (4)$$

where $\tau(S)$ denotes the logical value (1 or 0) of the statement $S$. The assumption made here is that each vehicle occupies the whole lane, so if ego vehicle center deviates from the middle of the lane, it is colliding with vehicle in the adjacent lane. It is important to note that this is different from driving in a lane when executing the plan, as the control is not ideal and the vehicle can deviate from the middle of the lane. Figure 2 shows an example of geometric representation of the vehicle obstacle within a defined search space $\Omega$. The presented vehicle speeds up and then slows down.

Beside collision, it is sometimes forbidden to overtake the vehicle from the right side. This can be expressed by a collision-test given in (5). This is modeled by prohibiting velocities greater than the velocity of a vehicle on the left. Beside using the velocity limit, position is used so that in the case when a vehicle tries to overtake the ego vehicle, and ceases overtaking for some reason, the ego vehicle does not slow down too. The corresponding collision test is formulated via:

$$\tau\left(\exists \mathbf{q} \in \mathcal{O}_k^V \mid l(t) \in [1, l_k(t)-1], \tfrac{\partial s(t)}{\partial t} \geq \tfrac{\partial s_k(t)}{\partial t}, s_k(t) \geq s(t)\right). \quad (5)$$

Practically, overtaking a vehicle requires only a velocity greater than the velocity of a vehicle. However, to limit the time of the overtaking maneuver, in several countries (e.g. Austria) there is also a limit on the minimum velocity difference $\Delta v_{ov}$, when overtaking other vehicles. The corresponding collision test can be formulated as:

$$\tau\left(\exists \mathbf{q} \in \mathcal{O}_k^V \mid l(t) \in [l_k(t)+1, N_l], \tfrac{\partial s(t)}{\partial t} \leq \tfrac{\partial s_k(t)}{\partial t} + \Delta v_{ov}\right). \quad (6)$$

In multilane urban driving scenarios, rules for overtaking are not applicable.

*2) Traffic light obstacle:* The traffic light is a traffic control device which prohibits passing the defined line, during specific periods in time on certain lane. Figure 3 shows obstacles created by two traffic lights across all lanes although they can be active on a single or several lanes. The obstacle is defined as:

$$\mathcal{O}_{ki}^{TL} = \{\mathbf{q} \in \Omega \mid s = s_k, l = l_k, t \in [t_{ki}, t_{ki} + \Delta t_{ki}]\}. \quad (7)$$

Each traffic light represents infinitely many obstacles, periodic in $t$, with the constant or variable period depending on the traffic light control system. The collision check is performed by evaluating:

$$\tau\left(\mathbf{q} \cap \mathcal{O}_{ki}^{TL} \neq \varnothing\right). \quad (8)$$

The vehicle trajectory should not pass trough the region of the traffic light at the time when the red light is on.

*3) Speed limit:* Speed limits may originate from speed limit signs, road curvature or some other factors. They are defined on certain segment of the road and active in the following region of $\Omega$:

$$\mathcal{O}_k^{SL} = \{\mathbf{q} \in \Omega \mid s \in [s_k, s_k + \Delta s_k]\}, \quad (9)$$

The collision check is validated by:

$$\tau\left(\exists \mathbf{q} \in \mathcal{O}_k^{SL} \mid \tfrac{\partial s(t)}{\partial t} \geq v_{k_{MAX}}\right). \quad (10)$$

On this segment, the vehicle velocity, represented by a gradient in direction of $t$, must not exceed the defined value.

*4) Forbidden lane-change, solid line:* Lane-change prohibition can be also defined on certain segments of the road. It is usually marked with the solid lane line. The obstacle representation is given as:

$$\mathcal{O}_{ki}^{LC} = \{\mathbf{q} \in \Omega \mid s \in [s_k, s_k + \Delta s_k], l(t) \in (l_i, l_i + 1)\}. \quad (11)$$

Prohibition may be applicable to both directions, where the collision check is performed by:

$$\tau\left(\mathbf{q} \cap \mathcal{O}_{ki}^{LC} \neq \varnothing\right). \tag{12}$$

Alternatively, the prohibition can hold for single direction. Left-wise lane change prohibition is defined via collision test (13), while the right-wise is defined via negative partial derivative.

$$\tau\left(\exists \mathbf{q} \in \mathcal{O}_{ki}^{LC} \mid \frac{\partial l(t)}{\partial t} > 0\right). \tag{13}$$

Obstacles formulated above are the most common constraints in everyday driving, and the majority of situations can be described by the combination of these. Clearly, multiple obstacles can be active at the same time.

It is worth pointing out that the collision checking with respect to such defined obstacles appears to be rather trivial, since it usually reduces to closed-form analysis whether some elementary, analytically defined curves intersect or not, or if the gradient of these curves attain certain values.

### D. Vehicle model

To model the vehicle motion for planning purposes, longitudinal and lateral dynamics should be derived. Assuming that the vehicle orientation does not deviate much from the road direction, the longitudinal motion is given by:

$$v(t) = \frac{\partial s(t)}{\partial t}, \tag{14}$$

$$a(t) = \frac{\partial v(t)}{\partial t} = \frac{F_m(t) - F_r(t)}{m}, \tag{15}$$

where $v(t)$ and $a(t)$ are velocity and acceleration along $s$, and $F_m(t)$ and $F_r(t)$ are forces generated by the motor and resistive force respectively. The rest of the vehicle model is presented in [25] where resistive forces and powertrain losses are modeled in detail. The vehicle model is used for computing the cost of a transition between certain states, $costtrans(v_i, v_f, t_t)$.

Since planning includes lane changes as well, modeling the lateral motion is of particular importance. This is not straightforward because of the vehicle kinematics and dynamics. For planning purposes, it is important that the model is conservative so that resulting trajectory is feasible, yet not too conservative to disregard many feasible trajectories. Therefore, the lateral motion is modeled as linear in time such that the vehicle needs a specific time $T_{LC}$ to execute the full lane change. This is consistent with [26], where the authors stated that most of the lane-changes are executed in 3-8 s. This simplification limits the use of a lane change on smaller velocities, which is acceptable as it can be considered a parking maneuver. Alternatively, the clearance for the lane change on smaller velocities can be provided with by increasing safety buffer around the obstacles.

### E. Cost function

A cost function is used to evaluate quality of a given trajectory. It can reflect multiple goals such as: short travel time [24], comfort [27], safety [28], energy efficiency [25], traffic

---

**Algorithm 1:** A* search for horizon

**input** : $n_{start}$, Obstacles data ($\mathcal{O}$), $h(s,v)$
**output**: $v_{ref}, l_{ref}, t_{ref}$ trajectory for horizon

1 **begin**
2    $n, n_r \leftarrow n_{start}$ /* Start pose */
3    CLOSED $\leftarrow \varnothing$ /* list of closed nodes */
4    OPEN $\leftarrow n$ /* list of opened nodes */
5    **while** $n \in [0, S_{hor}] \times [0, T_{hor}]$ **and** OPEN $\neq \varnothing$ **and** !timeout **do**
6      CLOSED $\leftarrow$ CLOSED $\cup\, n$
7      OPEN $\leftarrow$ OPEN $\setminus n$
8      **foreach** $n' \in$ Expand$(n, \mathcal{O}, h(s,v))$ **do**
9        **if** $n' \in$ CLOSED **then**
10          *continue*
11        **else if** $n' \in$ OPEN **then**
12          **if** *new $n'$ is better* **then**
13            $n'$.parent $\leftarrow n$ /* update parent */
14          **else**
15            *continue*
16        **else**
17          OPEN $\leftarrow$ OPEN $\cup\, n'$ /* add to list */
18          **if** $n'$ *closer to horizons than* $n_r$ **then**
19            $n_r \leftarrow n'$
20      $n \leftarrow$ argmin $n.f \in$ OPEN
21    *reconstruct trajectory starting from $n_r$ backwards*
22    **return** *trajectory*

---

rules (driving on the rightmost lane [29]) or a combination of these [30]. The design of the cost function is particularly important as it influences vehicle behavior and defines the optimal solution. Cost function used in this work is reflecting energy efficiency and is explained in details in [25], [31] with the additional constant cost for each lane change. The cost function is kept simple while other desired behaviors are implicitly defined through the obstacle formulation.

## III. OPTIMAL MOTION PLANNING FRAMEWORK

In this section, we elaborate on the proposed optimal MP framework. First, some general aspects of the framework are described, followed by the clarification of individual features.

### A. Framework

The proposed framework is based on A* search method [32], guided by an exact cost-to-go map from a relaxed problem in an MPC-like replanning scheme. Each $T_{rep}$ seconds, replanning is triggered with the current measurement of positions and velocities of other vehicles, traffic lights timing data together with the map data. Based on measurements, the motion of other vehicles is predicted and collision-free trajectory for a defined horizon is generated.

The trajectory is generated by a grid-like searching using A*. The grid is constructed by the discretization of $t$, $s$ and $l$ from the original continuous search space definition. Velocity $v$ is additionally used to provide completeness because of the longitudinal dynamics of the model. Starting from the initial configuration, defined as the *initial node*, chosen as the first *current node*, all neighbors are determined by expanding the *current node*. The resulting *child nodes* are added to the OPEN list. If the *child node* is already in the OPEN list, and new *child node* has a lower cost, the parent of that node is updated, otherwise it is ignored. From the OPEN list, the node with the lowest cost is chosen to be the next *current*

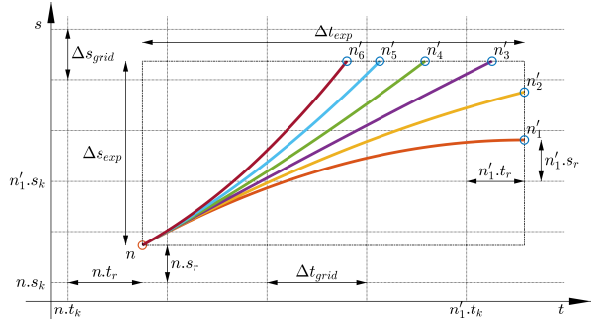

Fig. 4: Expanding parent node $n$ to different child nodes $n'$ by piecewise constant acceleration.

*node* and the procedure is repeated until horizon is reached, the whole graph is explored or the computation time limit for planning is reached. Finally, the node closest to the horizons is used to reconstruct the trajectory. The pseudocode for this procedure is presented in Algorithm 1.

To avoid rounding errors, as the expansion of node creates multiple transitions which in general do not end at gridpoints, the hybrid A* approach [9] is used for planning. Hybrid A* also uses the grid, but keeps continuous values for the next expansion without rounding it to the grid, thus preventing the accumulation of rounding errors.

As $v$ belongs to the discrete set of values as defined in the expansion (Algorithm 2), the hybrid A* approach is used only for $t$, $s$, $l$. Therefore, each *node* $n$ contains 14 values: four indices for $v$, $t$, $s$, $l$, ($n.v_k$, $n.t_k$, $n.s_k$, $n.l_k$), four indices for the *parent node* (to reconstruct trajectory), three remainders from the discretization of $t$, $s$ and $l$ ($n.t_r$, $n.s_r$, $n.l_r$), the direction of the lane-change $n.l_{dir}$, the exact cost-to-come to the node ($n.g$), and the estimated total cost of traveling from the initial node to the goal region ($n.f$). The value $n.f$ is computed as $n.g + h(n)$, where $h(n)$ is the heuristic function.

The planning clearly requires processing time. The compensation of the planning time can be achieved by introducing $T_{plan}$, a guaranteed upper bound on planning time. The planning is then initiated from a position where the vehicle would be after the $T_{plan}$. The old trajectory is executed while the new one is being processed. Thus, the new trajectory is already planed when $T_{plan}$ arrives. This approach has been widely used in MP for automated vehicles [11].

*B. Node expansion*

To build trajectories iteratively, *nodes* are expanded and *child nodes* are generated, progressing toward the goal. From each node $n$, only dynamically feasible and collision-free *child nodes* $n'$ should be generated. A single *child node* is generated for each possible longitudinal and lateral motion variant. The procedure is presented in Algorithm 2.

The longitudinal motion variants are generated by assuming uniform accelerations from the inherited parent velocity, so that the discrete final velocities (represented by the array $v_f$) are reached at expansion limits. Expansion limits are defined by $\Delta s_{exp}$ for distance and $\Delta t_{exp}$ for time (Figure 4). Since trajectories reflect the motion with the uniform acceleration, the average velocity of a specific motion variant

**Algorithm 2:** Expand function

**input** : $n$, Obstacles data ($\mathcal{O}$), $h(s,v)$
**output**: $n'$ array

1 **begin**
   /* generate array $n_{lon}$ of longit. variants */
2 $\quad v_i \leftarrow n.v_k \cdot \Delta v$
3 $\quad v_f \leftarrow [0 : \Delta v : v_{max}]$
4 $\quad n_{lon} \leftarrow$ transitions from $v_i$ to $v_f$
5 $\quad n_{lon}.g \leftarrow n.g + costtrans(vi, vf, t_t)$
6 $\quad n_{lon}.f \leftarrow n_{lon}.g + h(s, v_f)$
7 $\quad n' \leftarrow n_{lon}$
   /* generate lateral variants and add to $n'$ */
8 $\quad$ **if** $mod(n.l_k, 1) \neq 1$ **then**
9 $\quad\quad$ progress $n'.l_k$ /* lane change in progress */
10 $\quad\quad$ increase $n'.f$ and $n'.g$
11 $\quad$ **else**
12 $\quad\quad$ **if** $n.l_k > 1$ **then**
13 $\quad\quad\quad$ $n_r \leftarrow n_{lon}$ /* lane change right */
14 $\quad\quad\quad$ increase $n_r.l_k$
15 $\quad\quad\quad$ increase $n_r.f$ and $n_r.g$
16 $\quad\quad\quad$ $n' \leftarrow n' \cup n_r$
17 $\quad\quad$ **if** $n.l_k < N_l$ **then**
18 $\quad\quad\quad$ $n_l \leftarrow n_{lon}$ /* lane change left */
19 $\quad\quad\quad$ decrease $n_l.l_k$
20 $\quad\quad\quad$ increase $n_l.f$ and $n_l.g$
21 $\quad\quad\quad$ $n' \leftarrow n' \cup n_l$
22 $\quad n_o \leftarrow \{x \in n' \mid \tau(x, \mathcal{O}) = 1\}$ /* collision check */
23 $\quad n' \leftarrow n' \setminus n_o$
24 $\quad$ **return** $n'$

equals $\overline{v} = (v_i + v_f)/2$. If $\overline{v} < \Delta s_{exp}/\Delta t_{exp}$, the trajectory will end on time expansion limit $\Delta t = \Delta t_{exp}$. Otherwise, it will end on the distance expansion limit $\Delta s = \Delta s_{exp}$. Distance and time values ($\Delta s$ and $\Delta t$) for each variant are summed with the *parent node*'s remainders $n.s_r$ and $n.t_r$. Resulting sums are used to generate *child nodes* by increasing *parent node*'s indices ($n.s_k$ and $n.t_k$) with the quotient of division of sums with discretization steps of the grid ($\Delta s_{grid}$ and $\Delta t_{grid}$), and computing new remainders. For each *child node* from the array $n'$, costs are computed as well. Cost-to-come is inherited from the parent and increased by the cost of transition. Cost-to-go is provided by the heuristic function explained in the following subsection. The compliance with the vehicle's internal constraints (e.g. maximum acceleration) is checked and the nodes that violate these constraints are removed.

Generated longitudinal motion variants are then used for lateral motion expansion. If the parent node is in the middle of the lane ( $l$ is the integer), variants for possible lane change right and left, beside staying in the lane are generated. Generated variants are tripled, one set of longitudinal variants for each. The values $n'.l_k$ and $n'.l_r$ are increased or decreased for the lane changes left or right respectively. They are modified by the value $\Delta t/T_{LC}$ based on the travel time of the particular variant, and the defined lane change time $T_{LC}$. If the parent node is already in the process of lane change, lane change is progressed without generating other lateral motion variants (ln 9, Alg. 2). Finally, compliance with obstacles such as other vehicles, traffic lights, etc. is checked and all child nodes and motion variants that are not collision-free are removed (ln 22, 23, Alg. 2).

## C. Heuristic function

The heuristic function $h(n)$ is used to estimate the cost needed to travel from some node $n$ to the goal state (*cost-to-go*). As it is shown in [32], if the heuristic function is underestimating the exact cost to go, A* search provides the optimal trajectory. For the shortest path search, the usual heuristic function is the Euclidian distance. To find the energy optimal velocity trajectory, the heuristic function must underestimate the energy needed to drive to the goal. In this framework, the *cost-to-go* map resulting from the backward search for the relaxed problem is used as heuristic function. The *cost-to-go* computation phase is executed only once at the beginning of the trip, or if the goal is changed. The computation is performed by using backward dynamic programming (DP), starting from the goal state ($s$ and $v$), backward in $s$, as it was shown in [25]. In this phase, only time invariant constraints are considered (e.g. speed limits) with topological road profile and the vehicle model without considering time-varying constraints. The resulting *cost-to-go* map is an admissible heuristic, as other vehicles can prohibit certain regions of the state space, which may only increase the cost to travel from the initial state to the goal region. This is valid if platooning effects are neglected, as platooning can potentially reduce the airdrag effect (which is considered in the initial computation), and decrease the cost of travel, but for compact vehicles, this effect is usually negligible. Heuristic function $h(s,v)$ depends only on $s$ and $v$. Using a similar approach as in [21], the resulting heuristic function is applied based on *child node*'s $s$ and $v$ values, while neglecting $t$ and $l$ values.

## D. Search horizons

As it was noted in III-A, the planing is performed until any of the trajectories reaches time ($T_{hor}$) or distance ($S_{hor}$) search horizon. Slower driving trajectories will reach the time, while faster trajectories will reach the distance horizon. Thus, the unnecessary planning can be avoided. If only one is chosen (e.g., $T_{hor}$) other one could adopt a large even infinite value. The search horizons should not be confused with local expansion limits, which uses a similar principle, but represents atomic motion segments when building the whole trajectory.

## E. Vehicle motion prediction

Though it is required for prediction of potential collisions, the perfect knowledge of the future motion of other vehicles is not available in principle. A naive way to predict the motion is to assume that the vehicles will continue to drive with their current velocity and stay in the current lane. On the other hand, a motion planning framework should provide collision-free plans even if trajectories deviate from the predicted one and the environment perception system introduces estimation errors. Therefore, safety buffers are used to increase obstacle regions, and frequent replanning is executed. The approach introduced in this framework is that the most intuitive prediction of driving (constant velocity) is used for finding the optimal trajectory, but an additional safety mechanism ensures a collision-free plan even for the worst case error regarding the relative distance estimation. This is provided by adding a step-like safety buffer to the obstacle. The lower and upper bound of the vehicle obstacle can be defined as:

$$\underline{\hat{s}}_k(t) = \hat{s}_k(t_0) + \hat{v}_k(t_0)(t - t_0) - L_S - s_b(t), \quad (16a)$$
$$\overline{\hat{s}}_k(t) = \hat{s}_k(t_0) + \hat{v}_k(t_0)(t - t_0) + L_S + s_b(t), \quad (16b)$$
$$s_b(t) = \begin{cases} \Delta s_{max}, & t_0 \leq t < t_0 + T_{rep}, \\ 3 \cdot \Delta s_{max}, & t_0 + T_{rep} \leq t \leq t_0 + T_{hor}, \end{cases} \quad (16c)$$

where $\Delta s_{max}$ is the maximum error of the vehicle relative distance estimation. The safety buffer $s_b$ is increased after $T_{rep}$ (the next replanning instance) to maintain robustness, so that in the next re-planning instance, the vehicle always starts from the position that is collision-free according to a new safety buffer. This is visualized in the Figure 5, showing the worst case scenario. The estimation error is such that in the first planning instance, $\Delta s_{max}$ is positive, while in the second instance, it is negative. It can be seen that the safety buffer from the first planning instance ensures that the trajectory is outside of obstacle area for the time interval $[t_0, t_0 + T_{rep})$ and the trajectory is outside of the safety buffer from the second planning instance for the time interval $[t_0 + T_{rep}, t_0 + 2 \cdot T_{rep})$. This safety buffer provides a partial robustness for deviations from the predicted trajectory as well, but no guarantees can be provided.

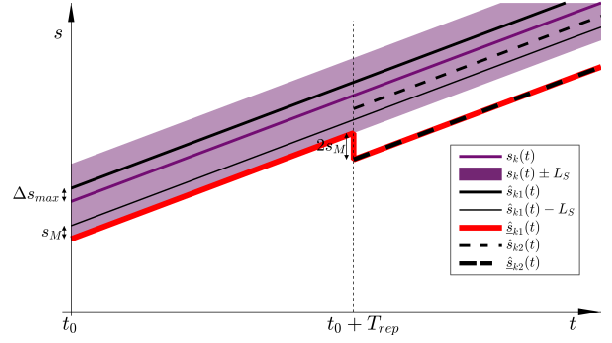

Fig. 5: Predicting movement of other vehicle - linearization.

## IV. SIMULATION STUDY AND DISCUSSION

The proposed MP framework was implemented in Matlab and used in Simulink as a Matlab compiled function together with PreScan, where a detailed vehicle model is simulated. Though, this implementation is not optimal, the results meet real-time requirements. For example, the proposed framework computes the plan with a horizons of $100m$ and $10s$, discretization of $\Delta v = 1m/s, \Delta s = 10m$, in **116 ms** in average. The computation time limits the maximum frequency of replanning to approx. $10Hz$.

For validation, three scenarios were created. The first scenario represents full stop due to road blockage to demonstrate ability of the proposed framework to provide slow trajectories. The second scenario represents lane change. These two scenarios were simulated with a detailed vehicle model. Due to space limits, these scenarios are only shown in the video[1]. The third scenario represents the urban multi-lane driving in a presence of traffic lights in dense traffic.

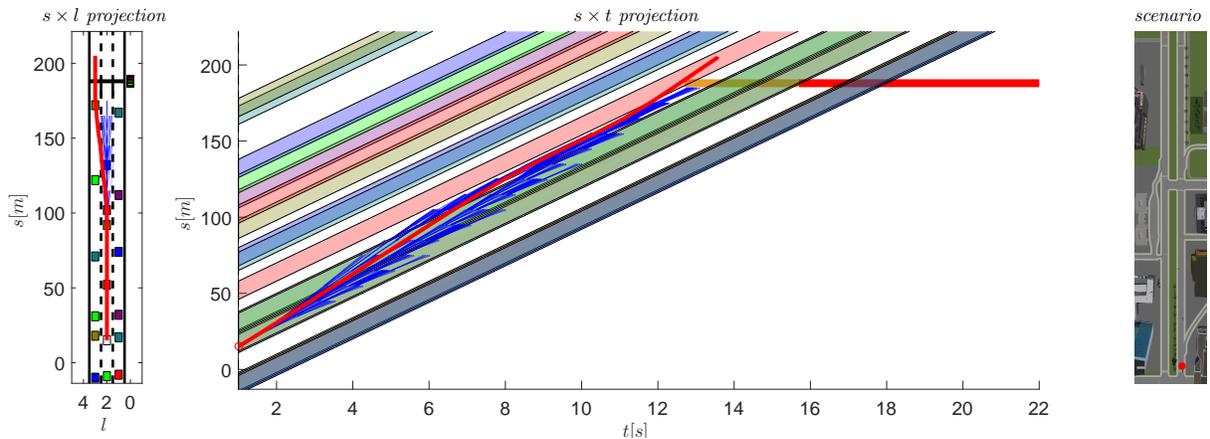

Fig. 6: The blue tree represents searched trajectories and red trajectory represents the final solution. Projections of trajectories on the $s \times l$ and $s \times t$ plane are shown on the left and middle plot respectively. Vehicles shown on the left plot represent polygon obstacles in the middle plot. The rightmost image is the screenshot from PreScan software showing the part of the real street used in the study. The resulting red trajectory shows vehicle reaching just behind red vehicle, slowing down to provide enough time for lane change left and speeding up to passing red vehicle while catching the green light.

*1) Detailed vehicle model:* The lane change feasibility within a sufficiently large time interval is validated using a higher fidelity model. The vehicle model used has 10 degrees of freedom (DoF) covering 6 DoF of the vehicle body and 4 DoF of vertical motion of unsprung masses. The vehicle body motion in space has longitudinal, lateral, vertical, roll, pitch and yaw motions. Assuming smooth driving in high friction conditions results in small wheel slip, the wheel rotational dynamics can be neglected. We assume the linear lateral characteristic of the tire, with the relaxation behavior included. The longitudinal motion of the vehicle body is modeled taking into account the applied wheel torques (both traction and brake torques), air drag, road resistance and slope forces.

*2) Urban scenario:* As for the third scenario, a segment of the street[2], approx. $750m$ long, was used. This segment contains three traffic lights, whose timings were experimentally obtained by observing a recorded video. Artificial traffic was created with the density of $30 veh/km/lane$ and the average velocity of $12 m/s$. All traffic participants follow the simple logic to satisfy traffic light signal, keep the current lane and keep appropriate spacing to others. Figure 6 presents one situation from the scenario. In this situation, the ego vehicle plans a lane change to pass the red vehicle in front and catch the green light. To make the clearance for lane change, the ego vehicle speeds up to get close to the preceding red vehicle (where the gap is), slows down during the lane change (to provide enough time for lane change) and again speeds up (to pass the red vehicle) to catch the green light. This situation truly demonstrates the importance of integrated planning for longitudinal and lateral motion.

To demonstrate the robustness, stochastic variations of the scenario are created by introducing randomized perturbations of initial positions and velocities. An uncompiled script version of the algorithm was used to facilitate the randomization of the scenarios and subsequent processing of the results. This results in somewhat larger runtimes than the previously shown. Nevertheless, this does not represent the culprit for validation since the cost and time of travel are the mayor indicators of the robustness in perturbed situations and computation time for compiled version is previously stated. For comparison, two different heuristics were used. The first one is the result of a DP approach from the relaxed problem [25], while the second one is a model-based one (MB) from [31]. Some numerical results of the simulation are shown in Table I. The results indicate that the proposed approach is robust to variations in the scenario and without significant deviations from the initial solution regarding the cost and time of travel. Moreover, using the DP heuristic is more effective compared to the MB heuristic, which is reflected in approx. **4 times** shorter **computation time** and the **number of nodes explored**. The variance of the computation time and the number of nodes explored are caused by variations in the complexity of driving situations.

[1] https://youtu.be/D5XJ5ncSuqQ
[2] The street "Zmaja od Bosne" in Sarajevo, Bosnia and Herzegovina, from "Trg Bosne i Hercegovine" to the Campus of the University of Sarajevo.

TABLE I: Comparison

| $h(n)$ | $T_{plan}[s]$ | nodes exp. | Cost $[kJ]$ | trav. time $[s]$ |
|---|---|---|---|---|
| DP | $1.39 \pm 0.75$ | $230 \pm 134$ | $405.8 \pm 20.5$ | $56.7 \pm 0.7$ |
| MB | $5.71 \pm 3.93$ | $1089 \pm 688$ | $416.3 \pm 11.7$ | $57.5 \pm 0.9$ |

## V. CONCLUSIONS

The proposed MP framework showed to be an efficient and robust solution for planning of automated driving, even in very complex scenarios such as multi-lane driving with traffic lights. The framework is capable of finding the velocity trajectory, such that enough clearance for lane-change is provided in tight situations, due to integrated longitudinal and lateral motion planning. We demonstrated the capability to provide slow and fast trajectories, which is particularly important for treating different constraints in urban driving. Moreover, using the exact DP cost-to-go as heuristic significantly improved efficiency of the search compared to MB heuristics.

Future work will explore several possible improvements: i) The search efficiency could be additionally improved by different variants of heuristic guided search algorithms; ii) Assuming sufficient computational resources, a more elaborated model that considers both road curvature and vehicle lateral dynamics could be used; iii) The improved approach should include more elaborated model of interaction with other traffic participants; iv) A more realistic perception system should imply probabilistic reasoning instead of assuming perfect knowledge about the environment.

## ACKNOWLEDGMENT

The project leading to this study has received funding from the European Unions Horizon 2020 research and innovation programme under the Marie Skodowska-Curie grant agreement No 675999, ITEAM project.

VIRTUAL VEHICLE Research Center is funded within the COMET Competence Centers for Excellent Technologies programme by the Austrian Federal Ministry for Transport, Innovation and Technology (BMVIT), the Federal Ministry of Science, Research and Economy (BMWFW), the Austrian Research Promotion Agency (FFG), the province of Styria and the Styrian Business Promotion Agency (SFG). The COMET programme is administrated by FFG.